%% file: main.tex
\documentclass[11pt]{article}

\usepackage[preprint]{acl}

\usepackage{times}
\usepackage{latexsym}

\usepackage[T1]{fontenc}
\usepackage{amsmath}
\usepackage{algorithmic}
\usepackage{amssymb}
\usepackage{algorithm}
\usepackage{listings}
\definecolor{lightgray}{rgb}{0.95, 0.95, 0.95}
\definecolor{darkgray}{rgb}{0.4, 0.4, 0.4}
\definecolor{backcolour}{rgb}{0.95,0.95,0.92}
\definecolor{myblue}{rgb}{0.2, 0.4, 0.8} 
\definecolor{mygreen}{rgb}{0.2, 0.6, 0.2} 
\usepackage{booktabs}

\usepackage[utf8]{inputenc}

\usepackage{microtype}

\usepackage{inconsolata}

\usepackage{graphicx}

\usepackage{array}
\usepackage{multicol}
\usepackage{multirow}
\newcolumntype{C}[1]{>{\centering\arraybackslash}m{#1}}

\usepackage{arydshln}

\usepackage{tikz}
\usepackage{xcolor}
\usepackage{pgfplots}
\pgfplotsset{compat=1.18}

\definecolor{monte_carlo}{RGB}{0,140,120}

\newcommand{\numwild}[2]{%
\begin{tikzpicture}[baseline]
    \pgfmathparse{#1 < #2 ? 1 : 0}
    \ifnum\pgfmathresult=1
        \pgfmathsetmacro{\percentdiff}{min(130, 130*(#2-#1)/#2)}
        \pgfmathsetmacro{\intensity}{\percentdiff}
        \fill[monte_carlo!\intensity!white, rounded corners=1]
            (-0.6em,-0.3em) rectangle (2.6em,1em);
    \fi
    \node[inner sep=0pt] at (1em,0.7ex) {#1};
\end{tikzpicture}%
}

\definecolor{monte_carlo}{RGB}{120,200,185}

\newcommand{\numheat}[1]{%
\begin{tikzpicture}[baseline]
    \pgfmathsetmacro{\intensity}{min(130, max(0, 130*#1))}
    \fill[monte_carlo!\intensity!white, rounded corners=1]
        (-0.6em,-0.3em) rectangle (2.6em,1em);
    \node[inner sep=0pt] at (1em,0.7ex) {#1};
\end{tikzpicture}%
}

\title{Cost-Awareness in Tree-Search LLM Planning: A Systematic Study}

\author{Zihao Zhang$^1$\ \ \ \ \ \ \ \ \ \  Hui Wei$^2$ \ \ \ \ \ \ \ \ \ \ \ Kenan Jiang$^1$  \ \ \ \ \ \ \ \ \ \  \\ \textbf{Shijia Pan$^2$ \ \ \ \ \  Kai Shu$^1$\ \ \ \ \ \ \ \ \ \ Fei Liu$^1$}\\
$^1$Emory University \ \ \ $^2$University of California, Merced \\
\texttt{\{zihao.zhang, kenan.jiang, kai.shu, fei.liu\}@emory.edu}, \\\texttt{\{huiwei2, span24\}@ucmerced.edu}
}

\begin{document}
\maketitle

\input{sections/1abstract}
\input{sections/2introduction}
\input{sections/3related_work}
\input{sections/4method}

\input{sections/5experiment}

\input{sections/6results}

\input{sections/7conclusion}


\bibliography{custom}
\appendix
\input{sections/9appendix}

\label{sec:appendix}


\end{document}

%% file: sections/1abstract.tex
\begin{abstract}

Planning under resource constraints is central to real-world decision making, yet most large language model (LLM) planners assume uniform action costs. 
We systematically analyze whether tree-search LLM planners are cost-aware and whether they efficiently generate budget-feasible plans. 
In contrast to black-box prompting, explicit search trees expose intermediate decisions, node evaluations, and failure modes, which allows for controlled ablations of planner behavior.
We study \emph{depth-first search}, \emph{breadth-first search}, \emph{Monte Carlo Tree Search}, and \emph{bidirectional search} within a unified framework. 
Our experiments show that existing tree-based LLM planners often struggle to find cost-optimal plans, and that additional search computation does not reliably improve optimality.
Among the methods evaluated, bidirectional search achieves the best overall efficiency and success rate. 
MCTS achieves the highest optimality on short-horizon tasks. 
Tree-search planners are especially valuable for studying LLM planning because their reasoning steps are explicit, in contrast to plain LLMs that internalize planning dynamics through post-training trajectories.
Our findings suggest that improving LLM planning under resource constraints will likely require new search algorithms, rather than solely scaling inference-time compute.

\end{abstract}

%% file: sections/2introduction.tex
\section{Introduction}\label{sec:intro}
Planning is a fundamental component of human intelligence, involving the generation of a sequence of actions to achieve a desired goal \cite{russell1995modern}. Recently, large language models (LLMs) have emerged as promising planners due to their broad world knowledge, strong reasoning capabilities, and demonstrated success across a wide range of complex domains \cite{huang2024understanding, wei2025plangenllmsmodernsurveyllm}, including task scheduling, e.g., travel and time management \cite{zheng2024natural, oh2025flextravelplannerbenchmarkflexibleplanning}, game playing \cite{wu2023smartplay, chen2023put, huang2024far, yang2025selfgoal, wang2023describe}, tool use \cite{wang2025actingreasoningmoreteaching}, and embodied decision-making \cite{ahn2022can, huang2022inner, zhou2024isr}. 

However, most existing LLM-based planners implicitly assume uniform action costs \cite{yao2022react, singh2023progprompt, sun2023adaplanner, prasad2024adapt}, treating all actions as equally expensive regardless of the resources they consume. In reality, actions often incur heterogeneous costs in terms of energy, time, or monetary expense, by an executor such as a human or a robot. This simplifying assumption significantly limits the applicability of LLM planners in real-world domains, where resource constraints are often critical. In practical settings (e.g., fire-fighting robots with limited battery capacity), ignoring heterogeneous action costs can lead to plans that exceed budget constraints, resulting in infeasible execution or even catastrophic failures.

\textbf{(RQ1)}. We investigate whether tree-search LLM planners such as Tree-of-Thoughts (ToT; \citet{yao2023tree}) and MCTS \citep{hao2023reasoning} are inherently \textbf{cost-aware}, that is, whether they account for heterogeneous action costs and explicit budget constraints during plan generation. We consider a broad class of search strategies, including depth-first search (DFS), breadth-first search (BFS), Monte Carlo Tree Search (MCTS; \citet{_wiechowski_2022}), and bidirectional search \cite{sadhukhan2012new,ren2024thinkingforwardbackwardeffective}, the last of which to the best of our knowledge has not been thoroughly explored in the LLM planning literature.

Compared to Input–Output (IO) and Chain-of-Thought (CoT) prompting \cite{wei2022chain}, tree-based search methods generally achieve superior planning performance due to their ability to explore a larger solution space, leverage search policies to guide exploration, and incorporate self-correction through backtracking or rollouts. These properties increase the likelihood of finding feasible and potentially optimal plans under complex constraints. Moreover, unlike classical cost-sensitive search algorithms such as A* search~\cite{su2025dualformercontrollablefastslow}, tree-search methods investigated in this study do not require task-specific reward or heuristic design, which simplifies their deployment and enables a more general and flexible planning framework across diverse domains.

\textbf{(RQ2).}\; We examine whether tree-search methods can \textbf{efficiently} generate cost-feasible plans under explicit budget constraints while requiring fewer tree-node expansions. This question is critical because tree-based methods are computationally expensive due to extensive node expansion and exploration, a challenge that is further amplified when LLMs are used to propose actions, evaluate states, and model state transitions.

We evaluate planner performance under three budget regimes with increasing flexibility:
(a) \textsc{Tight}, where the budget equals the cost of the ground-truth optimal plan;
(b) \textsc{Loose}, which allows a bounded cost overrun defined as the optimal cost plus a margin; solutions within this margin are considered acceptable even if they are not strictly optimal; and
(c) \textsc{Unlimited}, which imposes no budget constraint and accepts all valid plans.
For each LLM planner, we measure:
\textbf{Success rate}, defined as the proportion of tasks for which the planner generates a valid, executable plan that satisfies the specified budget regime;
\textbf{Optimality}, how close the cost of generated plans is to the minimum-cost optimal plan; and
\textbf{Search efficiency}, measured by the number of tree-node expansions required to find a budget-feasible plan.

Our study yields three main observations. First, tree-based LLM planners often struggle to identify cost-optimal plans. Second, increasing search computation does not consistently lead to improved optimality. Third, we compare multiple tree-search algorithms and find substantial differences in efficiency: bidirectional search emerges as the most efficient approach, while also achieving higher success rates on long-horizon tasks. Collectively, these findings highlight two promising directions for advancing cost-aware LLM planning: (a) learning cost-aware reward models to better guide node expansion, and (b) incorporating principled infeasibility pruning based on cumulative cost and lower bounds on remaining cost to eliminate branches that cannot satisfy budget constraints or improve upon the current best plan.

\begin{figure*}[t!]
\centering
\includegraphics[width=\textwidth]{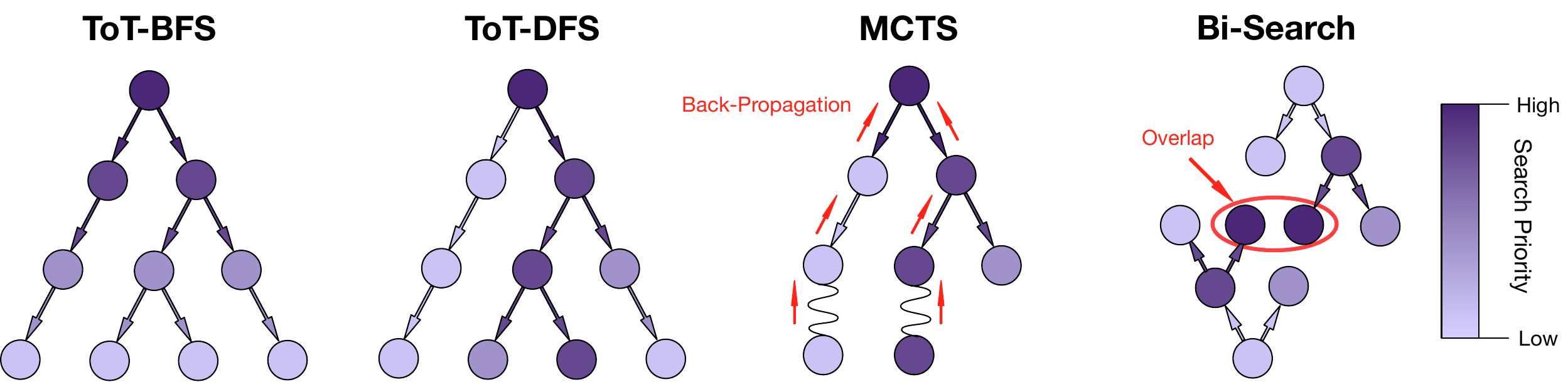}    
\label{fig:tree_search_example}
\vspace{-0.1in}
\caption{A visual comparison of four tree-based LLM planners included in this study, where node color intensity indicates search priority. \textbf{ToT-BFS} combines an LLM with breadth-first search, exhaustively exploring all nodes at each depth. \textbf{ToT-DFS} uses depth-first search, prioritizing deeper expansions during the search process. \textbf{MCTS} implements the Monte Carlo Tree Search (MCTS), combining exploration and previous roll-out by Upper Confidence Bounds for Trees (UCT equation~\ref{eq:ucb}) to dynamically guide the search process. \textbf{Bidirectional Search (Bi-Search)} incorporates two search trees, one from the initial state, one from the goal state, and alternates expansions to find a connecting state, thereby reducing effective search depth and pruning the search space.}
\label{fig:tree_search_example}
\vspace{-0.1in}
\end{figure*}

%% file: sections/3related_work.tex
\section{Related Work}

\paragraph{Tree-Search LLM Planners.}\;\;
Prior work on LLM-assisted planning has explored various tree-search paradigms, such as \emph{depth-first search (DFS)}, \emph{breadth-first search (BFS)}, and \emph{Monte Carlo Tree Search (MCTS)}.
BFS and DFS form the basis of Tree-of-Thoughts (ToT; Yao et al., 2023)\nocite{yao2023tree}, which enables systematic exploration and self-correction through intermediate reasoning states, with subsequent analyses examining bounded variants to balance solution quality and computational cost~\citep{katz2024thought}. MCTS-based planning methods integrate LLMs as policy, value, or world models to balance exploration and exploitation via rollouts, achieving improved robustness and generalization on complex planning tasks~\citep{_wiechowski_2022, hao2023reasoning, zhou2024languageagenttreesearch, zhang2025spiralsymbolicllmplanning}. Moreover, implicit search shifts the burden of exploration from inference time to training time. During post-training, the LLM learns to generate action sequences that yield high rewards, effectively internalizing a search policy over a latent space of trajectories~\citep{zhou2024languageagenttreesearch,parmar2025plantuningposttraininglanguagemodels}. 


However, existing approaches primarily optimize for task success or reasoning quality and largely ignore heterogeneous action costs and explicit budget constraints. In contrast, our work systematically studies the cost awareness and search efficiency of LLM-based tree-search planners, evaluating their ability to generate cost-feasible plans under budget constraints while minimizing tree-node expansions.

\paragraph{Cost-Aware Planning.} \;\;
Planning is a core component of human intelligence, centered on generating action sequences to explore solutions and support decision-making \citep{hayes1979cognitive, mattar2022planning, su2023language}. Much of the prior work emphasizes planning completeness, i.e., whether a planner can find a feasible plan when one exists \citep{puig2018virtualhome, shridhar2020alfworld, valmeekam2023planbench, li2025planetcollectionbenchmarksevaluating, parashar2025inferencetimecomputationsllmreasoning}. More recent studies incorporate action costs, making benchmarks more realistic by requiring solutions to balance feasibility with optimality \citep{wu2025catp, liu2025costbench, qian2025xroutertrainingcostawarellms}. TravelPlanner is a commonly used cost-aware benchmark, but it primarily evaluates executability: the objective is to produce a plan that satisfies all constraints rather than to minimize cost \citep{xie2024travelplanner}. In contrast, many other cost-aware benchmarks are difficult to scale in difficulty or to evaluate reliably, especially with respect to intermediate states and plan correctness. For these reasons, we use BlocksWorld as our testbed~\citep{kambhampati2024llmscantplanhelp}.


%% file: sections/4method.tex
\section{Methods}
We study cost-constrained planning problems where a planner is given an initial state $s_0$, a goal state $s_g$, a budget limit $\mathcal{B}$, and an action set $\mathcal{A}$. Executing an action $a \in \mathcal{A}$ in a state induces a transition to a successor state and incurs an action-dependent cost $c(a)$. The planner’s objective is to select an action sequence $\mathcal{A}' = [a_1, a_2, \dots, a_k]$ such that applying these actions from $s_0$ reaches the goal state $s_g$. In addition to goal reachability, the plan must satisfy the budget constraint, i.e., the total cost $\sum_{i=1}^k c(a_i)$ does not exceed $\mathcal{B}$.

\subsection{Tree-based LLM planner}
\label{sec: tree-search}
We study representative tree-search planners that integrate an LLM into the search loop, differing mainly in (1) how search policy is determined and (2) tree search structure, which determines whether the planner expands a single forward tree rooted at the initial state or two coupled trees rooted at the initial and goal states that terminate when the frontiers meet (Figure~\ref{fig:tree_search_example}). Across all methods, a search node represents a partial plan prefix $A_{partial}=[a_1,\ldots,a_t]$, with $t\in [1,K]$ where K is the maximum allowed number of actions per trajectory, and executing this prefix from $s_0$ yields the intermediate state $s_t$. Each node maintains the proposed action, state, the accumulated execution cost, and a scalar score used to prioritize further expansion (defined in \S\ref{sec:reward_design}). A node becomes terminal when it reaches the goal state or exceeds the maximum number of allowed actions per trajectory

\paragraph{Tree-of-Thought (BFS/DFS).} 
ToT explores the tree in a BFS/DFS-style manner by prompting the LLM to propose multiple candidate next actions from the current state $s_t$. At each expansion, we sample $m$ actions $\{a^{(j)}_{t+1}\}_{j=1}^{m}$, execute them to obtain successor states $\{s^{(j)}_{t+1}\}$, score each child with reward $r^{(j)}_{t+1}$, and select which nodes to continue expanding using either depth-first (DFS) or breadth-first (BFS) traversal

\paragraph{Monte Carlo Tree Search (MCTS).}
MCTS balances exploration and exploitation via iterative \textit{selection, expansion, simulation, backpropagation}. In the selection stage, starting from the root, we choose a leaf node from the current search tree that maximizes a UCT-style criterion:
\begin{equation}
\label{eq:ucb}
\text{UCT}(u) = Q(u) + \beta \sqrt{\frac{\ln N(\mathrm{parent}(u))}{1 + N(u)}},
\end{equation}
where $N(u)$ is the visit count and $Q(u)$ is the running estimate of node value. In the expansion stage, we query the LLM for $m$ candidate actions at the selected leaf node and add feasible children. In the simulation stage, we compute each new child’s reward $r$ (\S\ref{sec:reward_design}) and treat it as a rollout/value estimate. We keep running the roll-out greedily till the terminal state. In the backpropagation stage, we propagate the value upward, e.g., by updating $Q$ with running averages from terminal node to root node. Compared to ToT, MCTS tends to focus computation on a smaller set of promising branches while still occasionally exploring under-visited alternatives based on previous rollout results. MCTS outperforms BFS and DFS by dynamically balancing exploration and exploitation, focusing computation on high-value outcomes rather than exhaustively or rigidly traversing the search tree.

\paragraph{Bidirectional Search.}
Bidirectional search (Bi-Search) grows two trees simultaneously: a forward tree rooted at $s_0$ and a backward tree rooted at $s_g$. A forward node represents a partial plan prefix from $s_0$ to some $s_t$, while a backward node represents a partial \emph{reverse} plan from $s_g$ to an intermediate state (when reverse transitions are available) or, more generally, a goal-conditioned subgoal state proposed by the LLM. The algorithm alternates expanding the frontier that appears more promising (according to reward and/or remaining budget), and declares success when the two frontiers meet at a compatible intermediate state. The detailed pseudocode is included in Appendix~\ref{app: bidirectional search}. Compared to unidirectional tree search, bidirectional search significantly shrinks the search space by meeting in the middle. By replacing one deep expansion with two shallower ones, it avoids the rapid combinatorial blow-up that dominates long-horizon unidirectional search.

\subsection{Reward Design}
\label{sec:reward_design}
To guide general search-tree expansion, we assign each non-root node a reward score that reflects how promising it is for reaching a valid solution. In our setting, we adopt the reward design from RAP and apply it uniformly across all tree-search algorithms evaluated in our experiments~\cite{hao2023reasoning}. Based on RAP's implementation, the reward model consist of two component, action evaluation and self evaluation. We attached the detailed prompt in Appendix~\ref{app: prompt}. Concretely, we compute each node’s reward by combining two LLM-derived confidence signals—one assessing the selected action relative to available choices, and another self-checking the selected action under the budget constraint. We next describe how each component is computed and how its confidence signals are used to score nodes during search-tree expansion

\textbf{Confidence Reward.}
\textit{Action evaluation.} Consider a leaf node \(v\) with parent node \(v_p\). Let \(a\) be the action that transforms \(v_p\) into \(v\), and let \(c_a\) be its cost. The parent node has an accumulated cost \(c_{v_p}\). To assess the quality of \(a\), we prompt the LLM with the parent state \(v_p\), the available action set \(\mathcal{A}\), the budget limit \(\mathcal{B}\), and the accumulated cost \(c_{v_p}\). The LLM is asked to predict the best next action. Since \(a\) was the action taken, we use the log-probability that the LLM assigns to \(a\) as the confidence reward for node \(v\). \textit{Self-evaluation.} Given the action \(a\), the budget limit \(\mathcal{B}\), the parent node \(v_p\), and its cost \(c_{v_p}\), we prompt the LLM to assess whether \(a\) is a good choice by outputting either \textit{good} or \textit{bad}. We then use the log-probability of generating \textit{good} as an additional confidence signal for node \(v\). The final reward for node v is obtained by summing these two log-probability signals (action-evaluation and self-evaluation).

\subsection{Metrics}
We evaluate the performance using three metrics: \textbf{Success Rate}, \textbf{Optimality}, \textbf{Efficiency}. The \textbf{success rate} checks whether the planner finds a valid, cost-adherent plan within the node expansion limits. \textbf{Optimality} evaluates execution cost quality by comparing the generated plan’s cost to that of the ground-truth optimal plan. We use a ratio to make optimality comparable across instances with different cost scales and to reflect proportional suboptimality. \textbf{Efficiency} measures how much search compute a method expends before finding the first budget-feasible plan, operationalized as the unused fraction of a fixed node-expansion limit. A higher efficiency value means the planner reached a solution while consuming a smaller portion of its allotted node expansion. We include the detailed equation in the Appendix~\ref{app: eval metrics}. 

\begin{table*}
\setlength{\tabcolsep}{4.5pt}
\renewcommand{\arraystretch}{1.1}
  \centering
  \scriptsize
  \begin{tabular}{@{}cl
  m{0.65cm}<{\centering}m{0.65cm}<{\centering}m{0.65cm}<{\centering}m{0.65cm}<{\centering}m{0.65cm}<{\centering}m{0.65cm}<{\centering}m{0.65cm}<{\centering}
  m{0.65cm}<{\centering}m{0.65cm}<{\centering}m{0.65cm}<{\centering}m{0.65cm}<{\centering}m{0.65cm}<{\centering}m{0.65cm}<{\centering}m{0.65cm}<{\centering}
  @{}}
    & & \multicolumn{7}{c}{(Easy Tasks) $\leftarrow$ \textbf{\footnotesize Success Rate} $\rightarrow$ (Hard Tasks)} &
        \multicolumn{7}{c}{(Easy Tasks) $\leftarrow$ \textbf{\footnotesize Optimality} $\rightarrow$ (Hard Tasks)}\\
    \toprule
    & \textbf{Plan Length ($L$)} & $L$=2 & 4 & 6 & 8 & 10 & 12 & \textbf{SR} & $L$=2 & 4 & 6 & 8 & 10 & 12 & \textbf{Opt}\\
    & \textbf{Task Count}  & 15 & 58 & 99 & 138 & 154 & 183 & \textbf{Avg} & 15 & 58 & 99 & 138 & 154 & 183 & \textbf{Avg} \\

    \midrule
     \multirow{7}*{\rotatebox{90}{\textcolor{red}{\textsc{Tight}}}}
     ~&CoT w/ Qwen3
     & \numheat{0.6} & \numheat{0.08} & \numheat{0.00} & \numheat{0.00} & \numheat{0.00} & \numheat{0.00} & \textbf{0.11}
     & \numheat{1.00} & \numheat{1.00} & -- & -- & -- & -- & \textbf{0.33}\\
     ~&CoT w/ GPT4.1
     & \numheat{0.4} & \numheat{0.08} & \numheat{0.00} & \numheat{0.00} & \numheat{0.00} & \numheat{0.00} & \textbf{0.08}
     & \numheat{1.00} & \numheat{1.00} & -- & -- & -- & -- & \textbf{0.33}\\
     ~&CoT w/ Claude
     & \numheat{0.8} & \numheat{0.75} & \numheat{0.26} & \numheat{0.26} & \numheat{0.07} & \numheat{0.01} & \textbf{0.36}
     & \numheat{1.00} & \numheat{1.00} & \numheat{1.00} & \numheat{1.00} & \numheat{1.00} & \numheat{1.00} & \textcolor{red}{\textbf{1.00}}\\
     ~& ToT-BFS
     & \numheat{0.80} & \numheat{0.19} & \numheat{0.00} & \numheat{0.00} & \numheat{0.00} & \numheat{0.00} & \textbf{0.17}
     & \numheat{1.00} & \numheat{1.00} & -- & -- & -- & -- & \textbf{0.33}\\
     ~& ToT-DFS
     & \numheat{0.86} & \numheat{0.33} & \numheat{0.04} & \numheat{0.00} & \numheat{0.00} & \numheat{0.00} & \textbf{0.21}
     & \numheat{1.00} & \numheat{1.00} & \numheat{1.00} & -- & -- & -- & \textbf{0.50}\\
     ~& MCTS
     & \numheat{1.00} & \numheat{0.92} & \numheat{0.50} & \numheat{0.11} & \numheat{0.05} & \numheat{0.00} & \textbf{0.43}
     & \numheat{1.00} & \numheat{1.00} & \numheat{1.00} & \numheat{1.00} & \numheat{1.00} & -- & \textbf{0.83}\\
     ~&Bi-Search
     & \numheat{1.00} & \numheat{0.92} & \numheat{0.48} & \numheat{0.12} & \numheat{0.11} & \numheat{0.07} & \textcolor{red}{\textbf{0.45}}
     & \numheat{1.00} & \numheat{1.00} & \numheat{1.00} & \numheat{1.00} & \numheat{1.00} & \numheat{1.00} & \textcolor{red}{\textbf{1.00}}\\

    \midrule
      \multirow{7}*{\rotatebox{90}{\textcolor{blue}{\textsc{Loose}}}}
     ~&CoT w/ Qwen3
     & \numheat{0.6} & \numheat{0.16} & \numheat{0.00} & \numheat{0.00} & \numheat{0.00} & \numheat{0.00} & \textbf{0.13}
     & \numheat{1.00} & \numheat{0.67} & -- & -- & -- & -- & \textbf{0.28}\\
     ~&CoT w/ GPT4.1
     & \numheat{0.6} & \numheat{0.33} & \numheat{0.23} & \numheat{0.15} & \numheat{0.17} & \numheat{0.15} & \textbf{0.27}
     & \numheat{0.39} & \numheat{0.75} & \numheat{0.31} & \numheat{0.36} & \numheat{0.57} & \numheat{0.54} & \textbf{0.49}\\
     ~&CoT w/ Claude
     & \numheat{0.6} & \numheat{0.91} & \numheat{0.61} & \numheat{0.48} & \numheat{0.35} & \numheat{0.24} & \textbf{0.53}
     & \numheat{0.52} & \numheat{0.92} & \numheat{0.54} & \numheat{0.59} & \numheat{0.56} & \numheat{0.58} & \textbf{0.62}\\
     ~& ToT-BFS
     & \numheat{1.00} & \numheat{0.26} & \numheat{0.00} & \numheat{0.00} & \numheat{0.00} & \numheat{0.00} & \textbf{0.21}
     & \numheat{0.90} & \numheat{0.81} & -- & -- & -- & -- & \textbf{0.29}\\
     ~& ToT-DFS
     & \numheat{1.00} & \numheat{0.58} & \numheat{0.09} & \numheat{0.01} & \numheat{0.00} & \numheat{0.00} & \textbf{0.28}
     & \numheat{0.92} & \numheat{0.88} & \numheat{0.38} & \numheat{0.29} & -- & -- & \textbf{0.41}\\
     ~& MCTS
     & \numheat{1.00} & \numheat{1.00} & \numheat{0.85} & \numheat{0.39} & \numheat{0.11} & \numheat{0.01} & \textbf{0.56}
     & \numheat{1.00} & \numheat{0.93} & \numheat{0.58} & \numheat{0.49} & \numheat{0.51} & \numheat{0.48} & \textcolor{red}{\textbf{0.67}}\\
     ~& Bi-Search
     & \numheat{1.00} & \numheat{0.92} & \numheat{0.93} & \numheat{0.88} & \numheat{0.79} & \numheat{0.71} & \textcolor{red}{\textbf{0.87}}
     & \numheat{1.00} & \numheat{0.84} & \numheat{0.58} & \numheat{0.43} & \numheat{0.41} & \numheat{0.36} & \textbf{0.60}\\

    \midrule
     \multirow{7}*{\rotatebox{90}{\textcolor{purple}{\textsc{Unlimited}}}}
     ~&CoT w/ Qwen3
     & \numheat{0.6} & \numheat{0.16} & \numheat{0.00} & \numheat{0.00} & \numheat{0.00} & \numheat{0.00} & \textbf{0.13}
     & \numheat{1.00} & \numheat{0.67} & -- & -- & -- & -- & \textbf{0.28}\\
     ~&CoT w/ GPT4.1
     & \numheat{0.8} & \numheat{0.41} & \numheat{0.29} & \numheat{0.18} & \numheat{0.23} & \numheat{0.19} & \textbf{0.35}
     & \numheat{0.30} & \numheat{0.85} & \numheat{0.42} & \numheat{0.40} & \numheat{0.45} & \numheat{0.49} & \textbf{0.48}\\
     ~&CoT w/ Claude
     & \numheat{0.6} & \numheat{0.83} & \numheat{0.58} & \numheat{0.47} & \numheat{0.42} & \numheat{0.32} & \textbf{0.54}
     & \numheat{0.70} & \numheat{0.90} & \numheat{0.49} & \numheat{0.52} & \numheat{0.63} & \numheat{0.44} & \textbf{0.61}\\
     ~& ToT-BFS
     & \numheat{1.00} & \numheat{0.38} & \numheat{0.01} & \numheat{0.00} & \numheat{0.00} & \numheat{0.00} & \textbf{0.23}
     & \numheat{0.90} & \numheat{0.81} & \numheat{0.52} & -- & -- & -- & \textbf{0.37}\\
     ~& ToT-DFS
     & \numheat{1.00} & \numheat{0.67} & \numheat{0.12} & \numheat{0.00} & \numheat{0.00} & \numheat{0.00} & \textbf{0.30}
     & \numheat{0.90} & \numheat{0.78} & \numheat{0.36} & -- & -- & -- & \textbf{0.34}\\
     ~& MCTS
     & \numheat{1.00} & \numheat{1.00} & \numheat{0.86} & \numheat{0.55} & \numheat{0.20} & \numheat{0.05} & \textbf{0.61}
     & \numheat{1.00} & \numheat{0.93} & \numheat{0.58} & \numheat{0.44} & \numheat{0.52} & \numheat{0.42} & \textcolor{red}{\textbf{0.65}}\\
     ~& Bi-Search
     & \numheat{1.00} & \numheat{1.00} & \numheat{1.00} & \numheat{1.00} & \numheat{1.00} & \numheat{1.00} & \textcolor{red}{\textbf{1.00}}
     & \numheat{1.00} & \numheat{0.84} & \numheat{0.60} & \numheat{0.43} & \numheat{0.40} & \numheat{0.32} & \textbf{0.60} \\

    \bottomrule
  \end{tabular}
  \vspace{-5pt}
  \caption{\textbf{Success rate and optimality by plan lengths under budget constraints.}
    Columns correspond to task subsets grouped by ground-truth plan length $L$ (shown with task counts $n$ in the header; shorter plans are easier).
    Left block reports \textbf{Success Rate} (fraction of tasks solved; $\uparrow$ better).
    Right block reports \textbf{Optimality} computed on \emph{successful} tasks only (e.g., Cost(Optimal plan) / Cost(Generated plan); $\uparrow$ better); entries are `\texttt{--}' when no instance is solved for that length. Under TIGHT budgets, any plan that is considered valid must match the optimal (minimum) cost, so optimality is trivially 1.00 whenever a method succeeds within the node limit.}
    \label{tab:overall}
\end{table*}

%% file: sections/5experiment.tex
\section{Experiments}
\label{sec: experiments}
\subsection{Datasets}
We introduce our testbed \textbf{Budget-BlocksWorld}, a cost-augmented version of  BlocksWorld benchmark~\citep{valmeekam2022large}. BlocksWorld is a widely used planning benchmark with several distinct blocks on a table. The objective of this task is to rearrange the blocks on the table from its initial state to the goal state using four deterministic actions: pick-up, put-down, stack, and unstack. We adopt BlocksWorld with action costs to transform it into Budget-BlocksWorld. We assign non-uniform costs to actions, making some actions (e.g., put-down/pick-up) more expensive than others (e.g., stack/unstack), which forces planners to trade off plan length against execution cost. Building on top of this setup, Budget-BlocksWorld obtains 1,008 tasks with 6-block instances, and categorizes the difficulty by the optimal plan length (minimum number of actions from initial state to goal), where longer horizons require reasoning over more complex state transitions and larger search space. 

\textbf{Ground Truth Generation.} To establish the cost distribution for our experiments, we evaluate several cost schedules and recompute ground-truth plans to ensure cost optimality. Using exhaustive search, we enumerate all possible plans and select the lowest-cost plan as the new ground truth.

\begin{figure}
\centering
\includegraphics[width=0.48\textwidth]{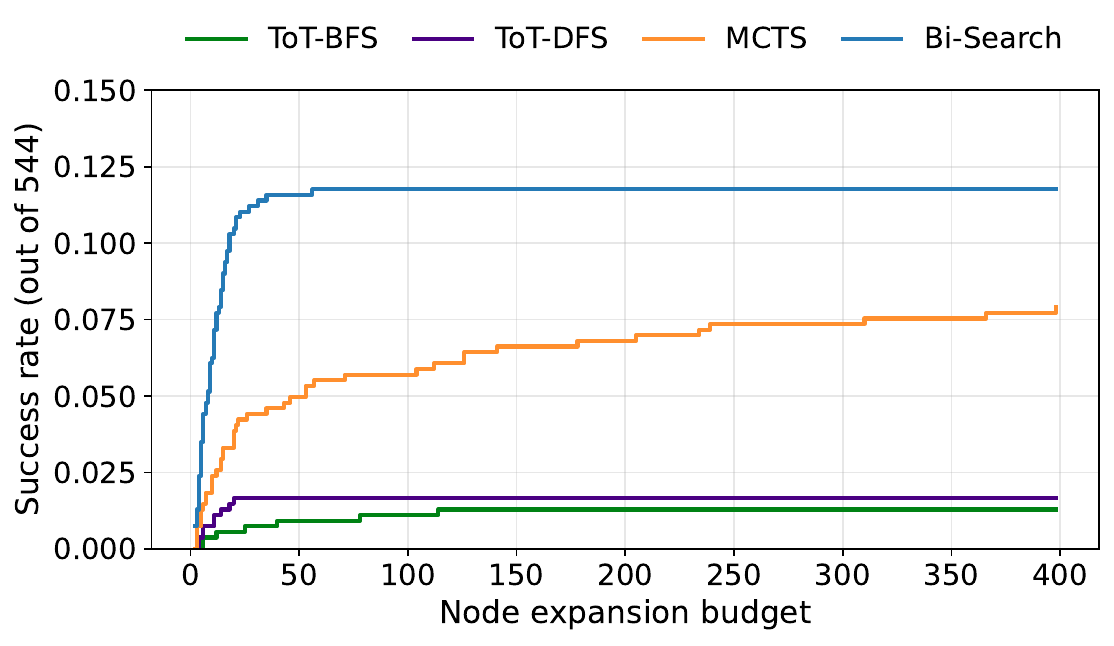}      
\vspace{-0.1in}
\caption{Node expansion vs Success rate. This plot demonstrates the number of node usage to find the optimal plan under \textbf{\textsc{TIGHT}} constraint.}
\label{fig:node_vs_sr}
\vspace{-0.1in}
\end{figure}

\textbf{Solution Shift Rate.} To ensure that our optimal solutions are clearly distinguishable from those of the vanilla BlocksWorld, we randomly select the task from the top-3 configurations with the highest ratio of shifted plans as our testbed.

A task's solution is considered as shifted if the cost-optimal ground truth plan is not identical to the uniform ground truth plan. We compute the solution shift rate for each cost schedule. This metric measures the proportion of tasks in Budget-BlocksWorld whose cost-optimal plans differ from the uniform-cost BlocksWorld baseline.  We provide more statistics on our cost-augmented BlocksWorld in Appendix~\ref{app: budget blocksworld}.


\textbf{Budget regimes.} Budget-BlocksWorld has three budget conditions to simulate different levels of cost strictness. \textbf{\textsc{Tight}} budget is set exactly to the ground-truth cost-optimal plan, permitting only plans whose cost matches the optimum. \textbf{\textsc{Loose}} budget is set to the optimal cost plus a small margin. More specifically, we set the margin to $2c_{max}$, 
where $c_{max}$ denotes the maximum cost among all actions. In BlocksWorld, plans typically consist of paired actions (e.g., pick-up/unstack followed by put-down/stack). Thus, setting the margin to $2c_{max}$ 
allows additional one high-cost action pair in the generated plan. \textbf{\textsc{Unlimited}} condition applies no budget constraints, and all valid plans are accepted. This setup allows us to systematically study planner's behavior under strict, relaxed, and unconstrained cost settings.

\subsection{Implementation details}
We compare a set of representative LLM planning approaches spanning (1) direct (raw) LLM planners with CoT that generate plans without explicit search (e.g., GPT-4.1 and Claude-Opus-4.1) and (2) tree-search-based planners, Tree-of-Thought (ToT-BFS/DFS), MCTS, and bidirectional search with Qwen3-8B as the backbone model. We adopt Qwen3-8B for tree-based planners for two reasons. First, tree search requires a large number of model calls, making closed-source models prohibitively expensive under our compute budget. Second, our reward computation relies on token-level scoring. 

\textbf{LLM planner w/ CoT.} We directly prompt the model with in-context learning examples and require it to generate the entire plan in one pass. The cost budget limit specified in the prompt is varied according to the given constraint. The full prompting details are included in the Appendix~\ref{app: prompt}. 



\textbf{Tree-based LLM planners.} We evaluate tree-search prompting approaches describe in Section~\ref{sec: tree-search}. For consistency, all the tree-based LLM planners use the same LLM-based confidence scoring signal described in Section~\ref{sec:reward_design} to rank candidate continuations. For each state, the LLM picks the top-5 actions from the action pool. To reduce hallucinated state transitions and ensure a shared notion of validity across methods, we use the PDDLGym~\citep{silver2016mastering} as the transit function to execute all proposed actions that deterministically update states and reject invalid actions.

\textbf{Feasibility pruning and search computation limits.} To assess whether cost-aware behavior arises from the model and search procedure \emph{without} relying on an explicit rule, we consider an optional \emph{hard budget feasibility filter} for ToT-BFS/DFS, RAP, and Bidirectional Search: any action (or partial plan) whose accumulated cost exceeds the budget is pruned immediately. We impose a common search budget of at most 500 node expansions for ToT-BFS/DFS, RAP, and Bidirectional Search to ensure comparable compute across methods and to prevent degenerate exhaustive search.

%% file: sections/6results.tex
\section{Experimental Results}
\subsection{Overall performance}

\begin{table}
\setlength{\tabcolsep}{4.5pt}
\renewcommand{\arraystretch}{1.1}
  \centering
  \scriptsize
  \begin{tabular}{@{}clm{0.48cm}<{\centering}m{0.48cm}<{\centering}m{0.48cm}<{\centering}m{0.48cm}<{\centering}m{0.48cm}<{\centering}m{0.48cm}<{\centering}m{0.6cm}<{\centering}}
    & & \multicolumn{6}{c}{(Easy Tasks) $\leftarrow$ \textbf{\footnotesize Efficiency} $\rightarrow$ (Hard Tasks)} & \\
    \toprule
    & \textbf{Plan Length} & $L$=2 & 4 & 6 & 8 & 10 & 12 & \textbf{Eff} \\
    & \textbf{Task Count}  & 15 & 58 & 99 & 138 & 154 & 183 & \textbf{Avg} \\
    \midrule
     \multirow{4}*{\rotatebox{90}{\textcolor{red}{\textsc{Tight}}}}
     ~& ToT-BFS  & \numheat{0.97} & \numheat{0.71} & -- & -- & -- & -- & \textbf{0.28} \\
     ~& ToT-DFS  & \numheat{0.99} & \numheat{0.96} & \numheat{0.86} & -- & -- & -- & \textbf{0.47} \\
     ~& MCTS     & \numheat{0.99} & \numheat{0.98} & \numheat{0.95} & \numheat{0.11} & \numheat{0.05} & -- & \textbf{0.51} \\
     ~& Bi-Search& \numheat{0.99} & \numheat{0.98} & \numheat{0.97} & \numheat{0.96} & \numheat{0.93} & \numheat{0.91} & \textcolor{red}{\textbf{0.96}} \\
     \midrule
     \multirow{4}*{\rotatebox{90}{\textcolor{blue}{\textsc{Loose}}}}
     ~& ToT-BFS  & \numheat{0.96} & \numheat{0.54} & \numheat{0.37} & -- & -- & -- & \textbf{0.31} \\
     ~& ToT-DFS  & \numheat{0.98} & \numheat{0.94} & \numheat{0.89} & \numheat{0.45} & -- & -- & \textbf{0.54} \\
     ~& MCTS     & \numheat{0.99} & \numheat{0.96} & \numheat{0.81} & \numheat{0.61} & \numheat{0.43} & \numheat{0.23} & \textbf{0.67} \\
     ~& Bi-Search& \numheat{0.99} & \numheat{0.99} & \numheat{0.98} & \numheat{0.97} & \numheat{0.93} & \numheat{0.91} & \textcolor{red}{\textbf{0.96}} \\
     \midrule
     \multirow{4}*{\rotatebox{90}{\textcolor{purple}{\textsc{Unlimited}}}}
     ~& ToT-BFS  & \numheat{0.95} & \numheat{0.78} & \numheat{0.35} & -- & -- & -- & \textbf{0.35} \\
     ~& ToT-DFS  & \numheat{0.98} & \numheat{0.92} & \numheat{0.83} & -- & -- & -- & \textbf{0.46} \\
     ~& MCTS     & \numheat{0.99} & \numheat{0.89} & \numheat{0.79} & \numheat{0.62} & \numheat{0.51} & \numheat{0.45} & \textbf{0.71} \\
     ~& Bi-Search& \numheat{0.99} & \numheat{0.98} & \numheat{0.96} & \numheat{0.94} & \numheat{0.91} & \numheat{0.87} & \textcolor{red}{\textbf{0.94}} \\
    \bottomrule
  \end{tabular}
  \vspace{-5pt}
  \caption{\textbf{Efficiency by plan lengths under budget constraints.} Efficiency is calculated by the ratio of unused node and node expansion limit ($\uparrow$ better; please refer to Eq. (\ref{eq: efficiency})). Same as Optimality, Efficiency is only computed on successful tasks. \textsc{`-'} means the algorithm does not find a feasible plan in the node expansion limit. }
  \label{tab:effciency}
\vspace{-0.1in}
\end{table}

\begin{figure*}[t!]
\centering
\includegraphics[width=1.0\textwidth]{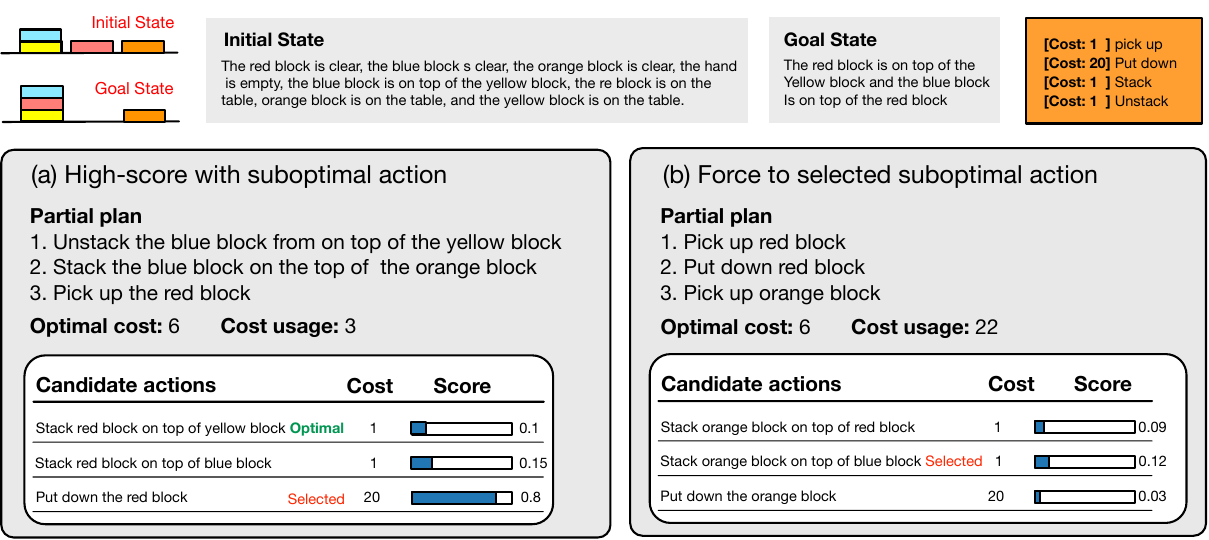}      
\vspace{-0.1in}
\caption{Decision-level diagnostics of cost-aware planning failures. We show the initial and goal states (top) and the action-cost schedule. Each panel zooms into a search step, listing the partial plan, cost usage versus the optimal cost, and the candidate actions with their model scores. (a) \textbf{Assign high score to suboptimal action}: the model assigns a high score to a costly suboptimal action (selected), diverting the search from the optimal continuation. (b) \textbf{Force to selected suboptimal action:} all candidates receive low, similar scores, forcing selection among suboptimal options and keep searching on the high cost branch.}
\label{fig:Example}
\vspace{-0.1in}
\end{figure*}



Table \ref{tab:overall} reports Success Rate and Optimality across ground-truth plan length $L=\{2,4,6,8,10,12\}$, under TIGHT, LOOSE, and UNLIMITED budget constraints, from which we draw the following insights: 


\textbf{Tree-based LLM planners struggle to find cost-optimal plans for long-horizon tasks.}
To illustrate these trends, we focus on the \textsc{Loose} and \textsc{Unlimited} budget settings, as optimality under the \textsc{Tight} constraint is uninformative: any successful plan must match the minimum cost, causing an optimality of 1.00 for all nonempty entries. Under both settings, Table~\ref{tab:overall} shows that tree-based methods (particularly MCTS and bidirectional search) achieve high optimality for shorter plan lengths ($L = \{2, 4\}$), where the planning problem is relatively easy. In these cases, MCTS outperforms closed-source LLMs using CoT prompting. However, as problem difficulty increases with longer horizons ($L = \{8, 10, 12\}$), the optimality of tree-based methods declines sharply and no longer surpasses strong closed-source baselines without tree search (e.g., CoT with Claude). 

Overall, these results suggest that: although tree search, especially bidirectional search, substantially improves \emph{feasibility} (i.e., success rate) over CoT-based methods using both open- and closed-source LLMs (shown in the left part of Table \ref{tab:overall}), node selection in existing tree-search planners is not consistently cost-aware, often producing cost-inefficient solutions even when valid plans are found in more challenging settings.

\textbf{More search compute does not necessarily yield optimal plan.} 
Figure~\ref{fig:node_vs_sr} shows optimality as a function of node expansions under the \textbf{\textsc{TIGHT}} budget constraint, where any successful plan is \emph{cost-optimal}. ToT-BFS/DFS and bidirectional search achieve most of their feasible solutions within the early stages of search, after which additional node expansions provide little benefit. MCTS continues to improve with increased search budget but quickly exhibits diminishing returns. This difference stems from MCTS, which updates node values based on rollout outcomes and provides more cost-aware guidance, whereas other tree-based planners rely on largely static search policies and can prematurely commit to suboptimal trajectories. Overall, these results show that additional computation alone does not ensure cost awareness, underscoring the need for explicit cost-aware guidance in LLM-based tree-search planning.


\textbf{Tree-search algorithm directly affects the efficiency.} 
Table~\ref{tab:effciency} shows large efficiency gaps across different tree-search strategies: bidirectional search can substantially improve the efficiency of finding feasible plans, even when the problem becomes more challenging. Moreover, Table \ref{tab:overall} demonstrates that compared with single-tree search methods, bidirectional search also achieves higher success rates on longer-horizon tasks. In contrast, \textit{ToT-BFS} tends to exhaustively expand nodes level by level; this layer-wise exploration quickly inflates the number of expansions, leading to poorer efficiency and a sharper drop in success as the horizon grows.

\subsection{Cost-Aware Failure Analysis}
In this section, we investigate failure cases in cost-aware planning to reveal the key limitations and challenges faced by tree-based LLM planners.

\subsubsection{Failure Modes Distribution}
\label{Sec: Outcome-level Failure}
\begin{figure}
\centering
\includegraphics[width=0.49\textwidth]{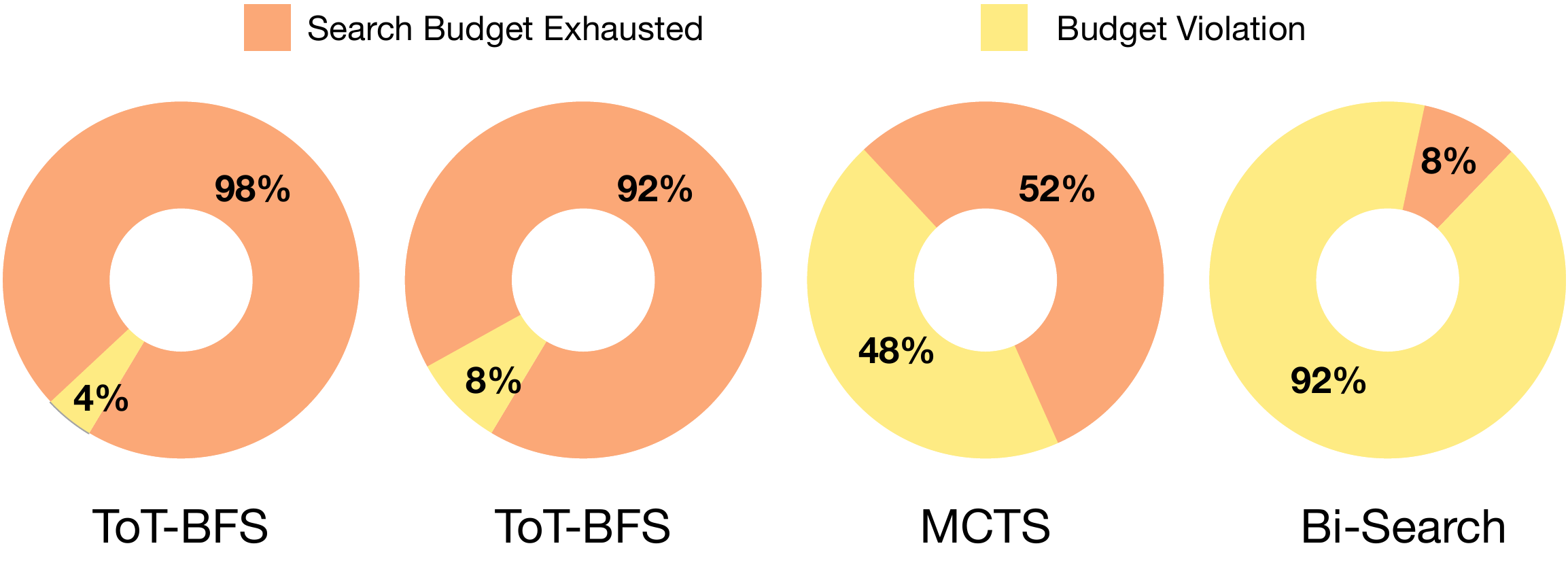}      
\vspace{-0.1in}
\caption{Failure mode distribution of \textsc{TIGHT} condition from $L=2$ to $L=8$}
\label{fig:Error}
\vspace{-0.1in}
\end{figure}

We categorize unsuccessful runs into two failure modes aligned with our research questions mentioned in Section \ref{sec:intro}. (1) \textbf{Budget Violation} denotes goal-reaching plans that exceed the cost budget, indicating limited cost awareness (RQ1). (2) \textbf{Search Exhaustion} denotes failures to find any valid plan within the node-expansion limit, reflecting poor search efficiency (RQ2).

\textbf{Overall Failure Distribution.} 
We conduct failure analysis under the \textsc{Tight} budget setting, as the \textsc{Loose} and \textsc{Unlimited} settings permit suboptimal plans and therefore do not meaningfully capture budget violations. Because ToT rarely finds feasible solutions for longer horizons (see Table \ref{tab:overall}), we restrict our analysis to tasks with plan lengths $L=\{2,4,6,8\}$.


Figure~\ref{fig:Error} presents the distribution of failure modes under the \textsc{Tight} budget across these plan lengths. We observe substantial differences across search methods: \textit{ToT-BFS} fails almost entirely due to Search Exhaustion (98\%), reflecting that its layer-wise breadth-first strategy cannot discover cost-optimal solutions within the node budget. \textit{ToT-DFS} exhibits a similar pattern (92\% Search Exhaustion, 8\% Budget Violation), with its depth-first commitment occasionally producing complete but cost-inefficient plans. \textit{MCTS} shows a more balanced failure distribution (52\% Search Exhaustion, 48\% Budget Violation); while it expands the search coverage compared to ToT methods, roughly half of its discovered solutions exceed the tight budget, revealing limited cost awareness. Finally, \textit{Bi-Search} experiences predominantly Budget Violations (92\%), suggesting that while bidirectional search efficiently finds feasible plans, it often fails to identify truly cost-optimal solutions.

\subsubsection{Root Cause Analysis}
\label{Sec: suboptimal-expansion}
\textbf{Search Trajectory Analysis and Failure Rate. } 
To further investigate why tree-based LLM planners violate budget constraints, we analyze their search traces through the lens of cost-optimal feasibility. For a node $s_v$ along a generated plan prefix, let $g(s_v)$ denote the accumulated cost from the initial state, and let $h(s_v)$ denote the \emph{optimal} remaining cost from $s_v$ to the goal, when such a path exists. Let $C$ be the \emph{ground-truth} optimal total cost. We say that the LLM fails to identify an infeasible trajectory at $s_v$ if $g(s_v) + h(s_v) > C$, meaning that even an optimal continuation from $s_v$ cannot satisfy the budget constraint. 

Based on this criterion, we define the \textbf{failure rate} as the proportion of nodes at which the LLM fails to recognize such infeasible trajectories. We randomly sampled 1000 nodes from the failure tasks. We report failure rates for different tree-based methods under the \textsc{TIGHT} and \textsc{LOOSE} budget constraints in Table~\ref{tab: error}, since the \textsc{UNLIMITED} setting accepts all feasible plans regardless of cost.  From Table~\ref{tab: error}, we derive the following insights, which highlight how different search algorithms perform under cost-aware planning:



\begin{table}[t]
\small
\centering
\begin{tabular}{lcccc}
\toprule
& \textbf{ToT-BFS}& \textbf{ToT-DFS} & \textbf{MCTS}  & \textbf{Bi-Search}\\
\midrule
\textsc{Tight} & 88.1\% & 89.3\% & 58.0\% & 77.3\% \\
\textsc{Loose} & 83.5\% & 85.7\% & 36.4\% & 69.8\% \\
\bottomrule 
\end{tabular}
\vspace{-6pt}
\caption{\label{tab:error_analysis}
Failure rate of identifying impossible trajectories ($\downarrow$ better). We evaluate 1000 nodes from $L=\{2, 4,6,8\}$ of the failure tasks. We describe the detailed definition of failure rate in \S\ref{Sec: suboptimal-expansion}.}
\label{tab: error}
\vspace{-16pt}
\end{table}


\textbf{LLM Fails to identify impossible trajectory.} Table~\ref{tab: error} shows that, even when explicitly prompted with action costs and cumulative cost, the LLM frequently fails to recognize prefixes that are already doomed under the budget, leading to budget-violating continuations in 36.4--89.3\% of node expansions. We observe two key insights from these results: (1) MCTS exhibits a substantially lower failure rate than the other search algorithms, suggesting that incorporating feedback from exploration can partially correct for miscalibrated confidence scores; and (2) holding the reward fixed, bidirectional search consistently outperforms ToT-BFS/DFS, implying that search structure and frontier management can reduce exposure to infeasible path even without changing the underlying reward signal. We discuss each finding in turn below.

\textbf{MCTS demonstrates notable advantages over heuristic-only approaches for cost-aware planning.} MCTS differs from the other search algorithms in how it selects nodes to expand: ToT-BFS/DFS and Bi-Search rely primarily on the LLM’s confidence score, whereas MCTS also uses feedback accumulated from prior exploration. As a result, MCTS achieves lower in failure rates than other method in both budget settings (Table~\ref{tab: error}). This stems from MCTS's dual guidance mechanism: in addition to heuristic scoring, it leverages value backpropagation to incorporate empirical evidence from rollouts, partially correcting for LLM scoring errors by learning which branches actually lead to budget-feasible solutions. 

\textbf{Tree-search algorithms matter under same reward signal.} 
Even using the same reward, failure rates are varied across BFS, DFS, and Bi-Search. In the DFS and BFS, when an LLM assigns uniformly low confidence to the available actions, the algorithm will still commit to a local choice and keep expanding an unpromising branch, shown on Figure~\ref{fig:Example}(b). While bidirectional search always selects the most promising branch on the search tree for both directions, this design can avoid getting trapped in the infeasible branches during the search process. Overall, these results show that the search strategy itself, not just the reward, critically determines whether the planner can avoid wasting computational resources on infeasible paths and improve the efficiency.

%% file: sections/7conclusion.tex
\section{Conclusion}
We analyze cost-aware, tree-based LLM planners under explicit budget constraints. Tree search substantially improves feasibility; however, increasing search compute does not reliably yield better cost-quality trade-offs (answer to RQ1). An analysis of search trajectories shows that the choice of search algorithm critically affects how efficiently a budget-feasible plan is found (answer to RQ2). Looking forward, we identify two promising directions: (i) learning cost-aware reward models to effectively guide node expansion, and (ii) introducing principled pruning via cost-aware lower bounds to eliminate branches unlikely to satisfy budget constraints or improve upon the current best plan.


\section{Limitations}
Our study intentionally focuses on tree-search planners, aiming to isolate how search structure alone affects budgeted planning; accordingly, we do not incorporate additional inference-time heuristics or self-refinement procedures that could confound this comparison. In addition, our results are conditioned on a specific prompting and LLM-based reward design. Finally, our experiments use a fixed set of LLM(s), and performance may vary in magnitude across other models even when the qualitative trends remain similar.

%% file: sections/9appendix.tex
\section{Budget-Blocksworld}
\label{app: budget blocksworld}

For each task in the BlocksWorld, we recompute the cost-optimal solution via exhaustive search. At each step, we expand all feasible actions and retain the minimum-cost path upon reaching a goal state. This non-trivial process ensures that our benchmark is grounded in reliable optimal solutions. Details are reported in Table~\ref{tab: budget_blocksworld}.

\begin{table}[p]
\setlength{\tabcolsep}{4.5pt}
\renewcommand{\arraystretch}{1}
\centering
\small
\begin{tabular}{lcccccc}
\toprule
\textbf{Action costs} & \textbf{L=2} & \textbf{L=4} & \textbf{L=6} & \textbf{L=8} & \textbf{L=10} & \textbf{L=12}\\
\midrule
$[20, 1, 1, 1]$ & 0\%  & 8\% & 33\% & 63\% & 64\% & 70\% \\
$[1, 20, 1, 1]$ & 0\% & 8\% & 32\% & 59\% & 59\% & 59\%  \\
$[1, 1, 20, 1]$ & 0\% & 16\% & 52\% & 74\% & 71\% & 78\% \\
$[1, 1, 1, 20]$ & 0\% & 0\% & 14\% & 39\% & 43\% & 52\%\\
$[20, 1, 20, 1]$ & 0\% & 0\% & 12\% & 35\% & 30\% & 35\%  \\
$[1, 20, 1, 20]$ & 0\% & 4\% & 14\% & 28\% & 26\% & 29\%  \\
$[20, 20, 1, 1]$ & 0\% & 4\% & 8\% & 16\% & 12\% & 18\%  \\
$[1, 1, 20, 20]$ & 0\% & 4\% & 8\% & 12\% & 32\% & 49\%  \\
$[20, 1, 1, 20]$ & 0\% & 8\% & 9\% & 37\% & 46\% & 54\%  \\
$[1, 20, 20, 1]$ & 0\% & 8\% & 40\% & 66\% & 63\% & 70\% \\
\bottomrule
\end{tabular}
\caption{\textbf{Solution shift rate under nonuniform action costs.} The tasks are classified by the length of optimal solution from L=2 to L=12. The first column represents the cost schedule allocated for each action: pick-up(pu), unstack(un), put-down(pd), stack(st). For example, [20, 1, 1, 1] means the cost of pick-up is 20, and the cost of other actions is 1. An optimal solution is counted as \emph{shifted} if its \emph{sequence of action types} under the given schedule is not exactly the same as under the uniform-cost solution. The rest columns represent the percentage of tasks whose optimal solutions are shifted.}
\label{tab: budget_blocksworld}
\end{table}

\section{Evaluation Metrics}
\label{app: eval metrics}
This appendix formally defines the metrics used in our experiments. \textbf{Optimality} measures solution quality by comparing the execution cost of a generated plan to the ground-truth optimal cost, with higher values indicating closer-to-optimal solutions. \textbf{Efficiency} measures computational cost in terms of node expansions, normalized by the fixed expansion budget; higher values indicate that a method finds a solution with fewer expansions.
\begin{align}
    \text{Optimality} = \frac{\text{Cost}_{opt}}{\text{Cost}_{gen}} 
    \label{eq: optimality}
\end{align}
\begin{align}
     \quad\text{Efficiency} = 1 - \frac{\#\,\mathrm{Nodes\ Expanded}}{\mathrm{Node\ Budget}}
    \label{eq: efficiency}
\end{align}

\section{Bidirectional Search Algorithm}
\label{app: bidirectional search}
Bidirectional search in our setting grows two complementary search trees to connect the initial and goal states more efficiently than a single forward tree. Bi-Search constructs a forward tree $T_f$ rooted at the initial state $s_0$ and a backward tree $T_b$ rooted at the goal state $s_g$, and expands them in alternating turns under a fixed node-expansion limit L. Each expansion selects a promising leaf node in the current tree, enumerates all feasible actions from that node (respecting the budget constraint), and adds the resulting successor states as children while tracking the cumulative path cost from the corresponding root. After each round, the algorithm checks whether the two trees contain an overlapping state; when an overlap is found, it terminates and extracts a complete plan by concatenating the forward path from $s_0$ to the overlap state with the backward path from the overlap state to $s_g$.  

To guide convergence, a confidence reward from the LLM (how promising the chosen action is given the state, remaining budget, and accumulated cost). This bidirectional structure reduces the depth each tree must explore before meeting, mitigating the exponential blow-up faced by single-tree search on long-horizon tasks. Algorithm \ref{alg: bidirectional} is the detailed pseudocode of bidirectional search. 
\begin{algorithm}[p]
    \caption{Bidirectional Search}
    \label{alg: bidirectional}
    \begin{algorithmic}[1]
    \REQUIRE Initial node $v_0 \in \mathcal{V}$, weight $\omega$,\\ goal node $v_g \in \mathcal{V}$, max iterations $L$, \\ budget limit $\mathcal{B}$
    \STATE Initialize confidence reward $R_{conf} : \mathcal{V} \to \mathbb{R}$
    \STATE Initialize feasible actions $A : \mathcal{V} \times \mathcal{B} \to \mathbb{A}$, \\child nodes $c : \mathcal{V}  \times \mathcal{A} \to \mathcal{V}$
    \STATE Initialize unvisited verifier $U: \mathcal{V} \to \{0,1\}$
    \STATE Initialize node set $\mathcal{V}_f \gets \{v_0\}$, $\mathcal{V}_b \gets \{v_g\}$
    \STATE Initialize empty plan $\pi \gets \emptyset$

    \FOR{$t \gets 0$ \TO $L - 1$}
        \STATE $v_f^* \gets \arg\max_{v \in Leaf(v_0)} \left(R_{conf}(v)\right)$
        \FOR{$a \in A(v_f^*, \mathcal{B})$}
            \STATE $v^* \gets c(v_f^*, a)$
            \STATE $\mathcal{V}_f \gets \mathcal{V}_f \cup \{v^*\}$
        \ENDFOR
        
        \STATE $v_b^* \gets \arg\max_{v \in Leaf(v_g)} \left(R_{conf}(v)\right)$
        \FOR{$a \in A(v_b^*, \mathcal{B})$}
            \STATE $v^* \gets c(v_b^*, a)$
            \STATE $\mathcal{V}_b \gets \mathcal{V}_b \cup \{v^*\}$
        \ENDFOR

        \STATE $v_{\text{overlap}} \gets \mathcal{V}_f \cap \mathcal{V}_b$
        \IF{$v_{\text{overlap}} \neq \emptyset$}
            \STATE $\pi \gets$ \textsc{ExtractPlan}$(v_{\text{overlap}})$
            \STATE \textbf{break}
        \ENDIF
    \ENDFOR
    \RETURN $\pi$
    \end{algorithmic}
    
\end{algorithm}

\section{Use of LLMs}
\label{sec:llm_use}
Large Language Models were used only to improve the clarity and presentation of the manuscript. They did not contribute to the research design, experiments, analysis, or conclusions, for which the authors take full responsibility.

\section{Prompting}
\label{app: prompt}

\noindent\textbf{Action Evaluation.}
We prompt the LLM with the current state, partial plan, candidate actions, goal state, and budget usage, and ask it to classify each action as good or bad. We use the log-likelihood of the \emph{good} label as a confidence for each action.

\begin{lstlisting}
I am playing with a set of blocks where I need to arrange the blocks into stacks. Here are the actions I can do and the corresponding time consumption: 
Pick up a block, 1 min
Unstack a block from on top of another block, 1 min 
Put down a block, 20 min 
Stack a block on top of another block, 1 min 
I have the following restrictions on my actions: 
I can only pick up or unstack one block at a time.  
I can only pick up or unstack a block if my hand is empty. 
I can only pick up a block if the block is on the table and the block is clear. A block is clear if the block has no other blocks on top of it and if the block is not picked up.  
I can only unstack a block from on top of another block if the block I am unstacking was really on top of the other block.  
I can only unstack a block from on top of another block if the block  
I am unstacking is clear. Once I pick up or unstack a block, I am holding the block.  
I can only put down a block that I am holding.  
I can only stack a block on top of another block if I am holding the block being stacked.  
I can only stack a block on top of another block if the block onto which I am stacking the block is clear.  
Once I put down or stack a block, my hand becomes empty.   

[STATEMENT] As initial conditions I have that, the red block is clear, the blue block is clear, the orange block is clear, the hand is empty, the red block is on the table, the blue block is on the table and the orange block is on the table. 
My goal is to have that the blue block is on top of the orange block. 
My previous plan is:  
[PREVIOUS PLAN]
pick up the blue block 
[PREVIOUS PLAN END]
The current time usage for the previous plan is 1 min  
My plan is as follows:   
[PLAN]
stack the blue block on top of the orange block 
[PLAN END]

As initial conditions I have that, <init state>\\
My goal is to <goals> 
My previous plan is:
[PREVIOUS PLAN]
<previous plan>
[PREVIOUS PLAN END] 
The current time usage for the previous plan is <current time usage> min
My plan is as follows: 
[PLAN]
\end{lstlisting}

\noindent\textbf{Self-Evaluated Confidence Reward}.
We prompt the LLM with the current state, partial plan, selected action, goal state, and budget usage, and ask it to classify the action as \emph{good} or \emph{bad}. The confidence score is computed as the log-likelihood assigned to the \emph{good} label.

\begin{lstlisting}
I am playing with a set of blocks where I need to arrange the blocks into stacks. Here are the actions I can do and the corresponding time consumption: 
Pick up a block, 1 min
Unstack a block from on top of another block, 1 min 
Put down a block, 20 min 
Stack a block on top of another block, 1 min 

I have the following restrictions on my actions: 
I can only pick up or unstack one block at a time.  
I can only pick up or unstack a block if my hand is empty. 
I can only pick up a block if the block is on the table and the block is clear. A block is clear if the block has no other blocks on top of it and if the block is not picked up.  
I can only unstack a block from on top of another block if the block I am unstacking was really on top of the other block.  
I can only unstack a block from on top of another block if the block  
I am unstacking is clear. Once I pick up or unstack a block, I am holding the block.  
I can only put down a block that I am holding.  
I can only stack a block on top of another block if I am holding the block being stacked.  
I can only stack a block on top of another block if the block onto which I am stacking the block is clear.  
Once I put down or stack a block, my hand becomes empty.  

[STATEMENT] The initial state: I have that, the yellow block is on the table. The red block is on the top of the yellow block. The green block is on the top of the red block. The blue block is on the green block. The blue block is clear. The hand is clear. 
The current state: I have that, the yellow block is on the table. The red block is on the top of the yellow block. The green block is on the top of the red block. The blue block is on the table. The blue block is clear. The hand is clear.
Goal: The green block is on the top of blue block.
[PREVIOUS ACTION]
unstack the blue block from the green block. 
put down the blue block.
We have used 21 minutes. 
Our time limit is 33 minutes 
[ACTION] unstack the green block from the red block 
[EVALUATION] good 

[STATEMENT] 
The initial state: I have that, <init state>
The current state: I have that, <current state>
Goal: <goal>
[PREVIOUS ACTION] <previous actions>
We have used <current time usage> minutes so far. 
Our time limit is <time limit> minutes.
[ACTION] <action>
[EVALUATION]
\end{lstlisting}

\noindent\textbf{Chain-of-Thought (CoT)} prompt used for GPT-4.1, Claude-Opus-4.1, and Qwen3-8B:

\begin{lstlisting}
 
I am playing with a set of blocks where I need to arrange the blocks into stacks. Here are the actions I can do and the corresponding time consumption: 
Pick up a block, 1 min
Unstack a block from on top of another block, 1 min 
Put down a block, 20 min 
Stack a block on top of another block, 1 min 

I have the following restrictions on my actions: 
I can only pick up or unstack one block at a time.  
I can only pick up or unstack a block if my hand is empty. 
I can only pick up a block if the block is on the table and the block is clear. A block is clear if the block has no other blocks on top of it and if the block is not picked up.  
I can only unstack a block from on top of another block if the block I am unstacking was really on top of the other block.  
I can only unstack a block from on top of another block if the block  
I am unstacking is clear. Once I pick up or unstack a block, I am holding the block.  
I can only put down a block that I am holding.  
I can only stack a block on top of another block if I am holding the block being stacked.  
I can only stack a block on top of another block if the block onto which I am stacking the block is clear.  
Once I put down or stack a block, my hand becomes empty.   

IMPORTANT: You must respond with ONLY the plan in the exact format shown below. Do not include any explanations, analysis, or additional text.

[STATEMENT] As initial conditions I have that, the red block is clear, the blue block is clear, the orange block is clear, the hand is empty, the red block is on the table, the blue block is on the table and the orange block is on the table.  
My goal is to have that the blue block is on top of the orange block.  
The time limit is 22 min
My plan is as follows: 
[PLAN]
pick up the blue block 
stack the blue block on top of the orange block 
[PLAN END]

[STATEMENT] As initial conditions I have that, the red block is clear, the blue block is clear, the orange block is clear, the hand is empty, the red block is on top of the yellow block, the blue block is on the table, the orange block is on the table and the yellow block is on the table. 
My goal is to have that the blue block is on top of the yellow block and the orange block is on top of the blue block. 
The time limit is 45 min
My plan is as follows: 

[PLAN]
unstack the red block from on top of the yellow block 
put down the red block 
pick up the blue block 
stack the blue block on top of the yellow block 
pick up the orange block 
stack the orange block on top of the blue block 
[PLAN END]

[STATEMENT] As initial conditions I have that, <init state>
My goal is to <goals>
The time limit is <time limit> min
My plan is as follows: 
[PLAN]

\end{lstlisting}